\documentclass[runningheads]{llncs}
\usepackage[T1]{fontenc}
\usepackage{amsmath}
\usepackage{amssymb}
\usepackage{graphicx}
\usepackage{multirow}
\usepackage{caption}

\begin{document}
\title{Matching in the Wild: Learning Anatomical Embeddings for Multi-Modality Images}
\titlerunning{Learning Anatomical Embeddings for Multi-Modality Images}

\author{Xiaoyu Bai\inst{1}  \and Fan Bai\inst{1}  \and Xiaofei Huo\inst{2} \and Jia Ge\inst{3} \and Tony C. W. Mok\inst{1}  \and Zi Li\inst{1}  \and Minfeng Xu\inst{1}  \and Jingren Zhou\inst{1}  \and Le Lu\inst{1}  \and Dakai Jin\inst{1}  \and Xianghua Ye\inst{3}  \and Jingjing Lu\inst{4} \and  Ke Yan\inst{1}}

\authorrunning{Xiaoyu Bai et al.}

\institute{DAMO Academy, Alibaba Group, Beijing, China\\
\email{bai.aa1234241@mail.nwpu.edu.cn}	\\
	\and United Family Hospital, Beijing, China
	\and Zhejiang University, Hangzhou , China
         \and Peking Union Medical College Hospital, Beijing, China
}



%
\maketitle              
\begin{abstract}
 Radiotherapists require accurate registration of MR/CT images to effectively use information from both modalities. In a typical registration pipeline, rigid or affine transformations are applied to roughly align the fixed and moving images before proceeding with the deformation step. While recent learning-based methods have shown promising results in the rigid/affine step, these methods often require images with similar field-of-view (FOV) for successful alignment. As a result, aligning images with different FOVs remains a challenging task. Self-supervised landmark detection methods like self-supervised Anatomical eMbedding (SAM) have emerged as a useful tool for mapping and cropping images to similar FOVs. However, these methods are currently limited to intra-modality use only. To address this limitation and enable cross-modality matching, we propose a new approach called Cross-SAM. Our approach utilizes a novel iterative process that alternates between embedding learning and CT-MRI registration. We start by applying aggressive contrast augmentation on both CT and MRI images to train a SAM model. We then use this SAM to identify corresponding regions on paired images using robust grid-points matching, followed by a point-set based affine/rigid registration, and a deformable fine-tuning step to produce registered paired images. We use these registered pairs to enhance the matching ability of SAM, which is then processed iteratively. We use the final model for cross-modality matching tasks. We evaluated our approach on two CT-MRI affine registration datasets and found that Cross-SAM achieved robust affine registration on both datasets, significantly outperforming other methods and achieving state-of-the-art performance.
\end{abstract}
\section{Introduction}
\begin{figure}[tb]
    \centering
    \includegraphics[scale=0.45]{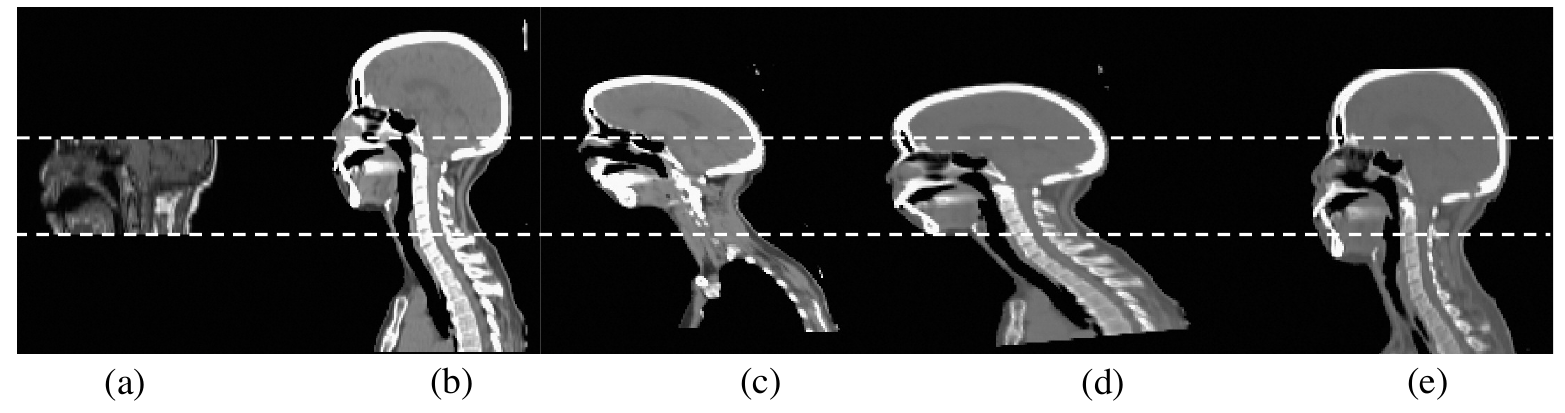}
    \captionsetup{skip=4pt}
    \caption{An example of one patient's CT and MRI scans with different FOV. (a) The original MRI with a small FOV. (b) The original CT with a large FOV. (c) CT to MRI affine registration result using the affine tool in DEEDS \cite{heinrich2013mrf}. (d) Affine result using C2FViT \cite{mok2022affine}. (e) Affine result of our method. We show CT to MRI results since MRI to CT results of~\cite{heinrich2013mrf,mok2022affine} have more severe distortion.}
    \label{fig:fig1}
    \vspace{-10pt}
\end{figure}

Medical images with multiple modalities are often used in clinical practice. For example, computed tomography (CT) and magnetic resonance imaging (MRI) are two widely used modalities capable of displaying detailed anatomy of the body. MRI is advantageous for localizing and characterizing lesions or organs due to its superior soft-tissue contrast. In contrast, CT offers lower soft-tissue contrast but higher resolution, providing information on tissue density and depicting the shape of anatomical structures. Accurate registration of CT-MRI images of the same patient is helpful for computer-aided diagnosis and precise radiotherapy treatment planning \cite{khoo2000comparison}. Both traditional~\cite{klein2009elastix,heinrich2013mrf} and deep-learning based~\cite{balakrishnan2019voxelmorph,mok2020large} registration methods have been successful in deformable registration, but they require the input fixed and moving images to be roughly aligned \cite{mok2022affine}. Therefore, a specific type of method exists that learns how to align two images using either rigid or affine transforms. Conventional approaches involve formulating affine registration as an iterative optimization problem within a predefined search space \cite{klein2009elastix,heinrich2013mrf}. Recently, there is a surge in learning-based methods \cite{huang2021coarse,mok2022affine,hoffmann2021synthmorph} which formulate affine registration as a regression task and using a differentiable spatial transform network (STN) to regress the affine matrix directly. However, all affine methods face challenges when aligning two images with significant differences in field-of-view (FOV). For instance, as illustrated in Fig. \ref{fig:fig1}, CT and MRI scans for head-and-neck tumors may have markedly different FOV in clinical practice. In such cases, both conventional and learning-based methods can become trapped in false local minima, resulting in misalignment. It is possible to manually crop the images to obtain a similar FOV before registration, but that will be tedious and not scalable.

One automated solution to this issue is to first perform key-point detection in both images \cite{o2018attaining,zhu2022datr}, and then match the identified key points and crop the two images to similar FOV. However, key-point detection methods typically require supervised training, which can increase the cost of label annotation (in multiple modalities) and limit their ability to identify landmarks that are not present in the training set. Recently, some methods have addressed the landmark detection problem by learning universal anatomical embeddings in a dense self-supervised manner. They can localize landmarks through template-query matching with only one template annotation \cite{yan2022sam,yao2021one,yao2022relative}. A representative method is SAM \cite{yan2022sam}, which aims to learn a distinct embedding for each voxel on the CT image to describe dense anatomical information, showing promising results on CT-to-CT point matching. SAM has been successfully applied to image registration~\cite{liu2021same}. The SAM embeddings can be used to compute correspondences between grid points on two images to estimate the affine transform matrix. Compared with other key-point-based methods, SAM-affine does not require any annotation to train and yields excellent results on cases with large FOV differences, making it an ideal option for quick and robust affine registration. However, SAM can only work with single-modality cases because its self-supervised design generates different embeddings for different appearances, but images of multiple modalities show large variations in appearance, especially in intensity distribution.

In this paper, we introduce Cross-SAM, a novel approach that produces the same embedding for the same anatomy in different modalities, enabling universal cross-modality point matching. A direct application of our method is the registration of multi-modality images (e.g. CT and MRI) with arbitrarily different FOVs. An iterative refinement process is designed that alternates between self-supervised embedding learning and unsupervised CT-MRI registration to learn Cross-SAM. 
We begin by applying aggressive intensity augmentation \cite{billot2023synthseg,billot2020learning} to both CT and MRI images, and then use both modalities (unpaired) to train an augmented SAM model. The aggressive data augmentation 
disrupts the intensity distribution of the two modalities, forcing the model to focus on higher-level structural information to describe anatomies, thus can learn approximately corresponding embeddings in CT and MRI. Consequently, we can apply the SAM-affine style registration to align the images of the same patient using rigid transform. 
While the rigid registration may not be highly accurate, it factors out the large linear misalignment between the image pair, making deformable registration~\cite{heinrich2013mrf} feasible and stable. 


The deformable registration result may still have distortions, but it roughly aligns the two images pixel-by-pixel. We subsequently train a cross-modality SAM using the registered data pairs to provide explicit cross-modality correspondence supervision. Compared to the initial augmented SAM model, the refined model exhibits superior cross-modality matching capabilities.  We can repeat the previous steps, carrying out SAM-affine and deformable registration and training a new cross-modality SAM, until the registration results stabilize and we obtain the latest cross-modality SAM model as our final model. 

We evaluate our method on two CT-MRI registration datasets of different body parts, head-and-neck and abdomen. They contain paired CT-MRI data of drastically different field-of-view. Our results demonstrate that our method can perform robust affine registration on both datasets and outperforms widely used traditional~\cite{heinrich2013mrf,klein2009elastix} and latest deep learning~\cite{mok2022affine} methods, achieving state-of-the-art performance. Our codes will be released upon acceptance.

\section{Method}
\begin{figure}[tp]
    \centering
    \includegraphics[scale=0.4]{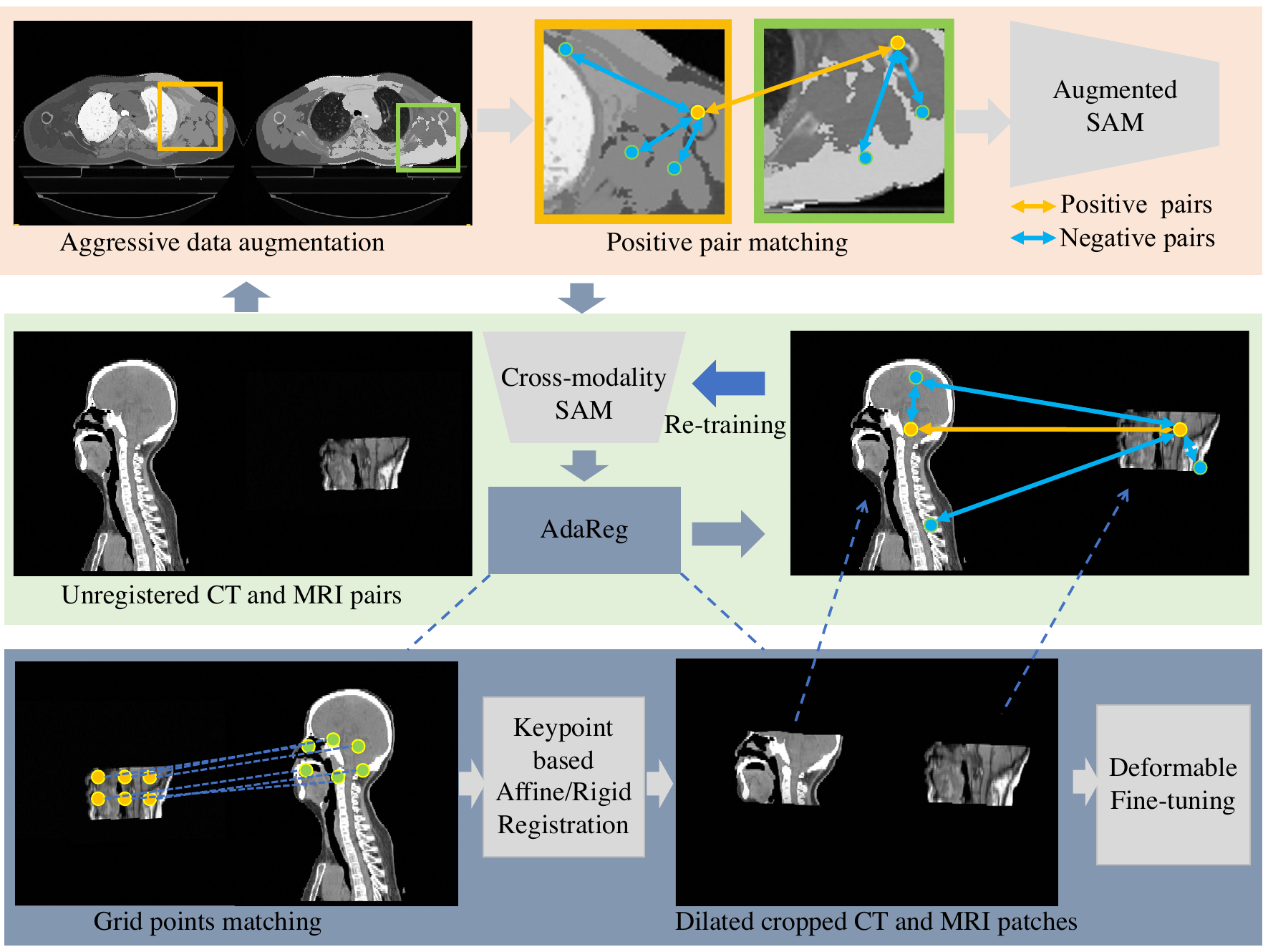}
    \caption{Diagram of the proposed method. Our method involves several steps. Firstly, we train a SAM using aggressive data augmentation. This module is then used to obtain voxel embeddings from unregistered CT-MRI pairs. Next, we utilize the grid points matching to estimate the affine/rigid transform matrix and align the two scans. To refine the alignment, we perform deformable fine-tuning by cropping the two scans to similar FOV. Due to the inaccuracy of the affine step, we retain a slightly larger FOV of the cropped scan. Finally, we use the registered paired data to retrain the SAM model, resulting in improved accuracy and performance.}
    \label{fig:framework}
\end{figure}
In this section, we present the details of the proposed Cross-SAM. It contains 3 main modules: (1) Augmented SAM; (2) Adaptive cross-modality registration (AdaReg); (3) Cross-modality SAM.

\subsection{Augmented SAM}
SAM (Self-supervised Anatomical eMbedding) \cite{yan2022sam} is a method that aims to create an anatomical embedding for each voxel on CT images. The goal is to ensure that similar body parts across different images have comparable embeddings. To achieve this, SAM employs a voxel-wise self-supervised learning approach. Specifically, SAM randomly crops two partially overlapped patches from a given CT image and applies random augmentation techniques. The overlapped region contains the same anatomical structure but appears differently on the two patches. SAM then uses the corresponding voxels on the two patches as positive pairs and treats the remaining voxels as negative samples to do contrastive learning \cite{oord2018representation}. The result is an embedding vector for each voxel. To match the same anatomical structure on different CT scans, SAM directly computes the cosine similarity of the template embedding vector to all the embeddings on the query image followed by the nearest neighbour (NN) matching.

SAM exhibits outstanding performance in CT-to-CT matching. However, it cannot be utilized for CT-MRI cross-modality matching due to the highly non-linear intensity discrepancy between the two modalities. Although style-transfer techniques \cite{yang2020unsupervised} may be feasible to generate pseudo CT from MRI, such methods often require large efforts to train and may struggle with different FOVs. Inspired by the modality-agnostic training concept in \cite{billot2023synthseg}, we train SAM with better modality robustness. We achieve this by applying strong or even aggressive contrast augmentation to both CT and MRI images. As shown in Fig. \ref{fig:framework}, we begin by using SLIC \cite{achanta2012slic} to segment an image into superpixels, and then apply random non-linear intensity transformation \cite{zhou2021models} and intensity inversion to each superpixel. This process generates images with visually unrealistic intensity, but it preserves the topology of anatomical structures. Furthermore, we include random affine transformation, noise, and blurring in our augmentation pipeline. We then select positive pairs from two aggressively augmented patches of the same image to train SAM. Thus, it is forced to learn features with less dependency with the intrinsic intensity distribution of particular imaging modality, and to focus on higher-level structural similarity such as organ layout, which is similar in CT and MRI. This augmented SAM can roughly match regions of the two modalities, providing a fully automated starting point for the following iterative refinement.
\vspace{-5pt}
\subsection{Adaptive Cross-modality Registration (AdaReg)} 
Using augmented SAM, we can find the correspondences of a set of points between the two modalities, and use them to estimate a linear transformation matrix between images. In SAM-affine \cite{liu2021same}, evenly spaced points were employed to calculate the affine matrix for two images. Since our CT-MRI pair is from the same individual, we favor computing the rigid transform matrix to prevent undesired shearing. Our task is thus to solve a least-square fitting problem of two 3-D point sets for rigid registration. Given two 3-D point sets ${p_i}$ and ${p_i'}$; $i=1,\cdots ,N$, where $p_i^{\prime}=R p_i+T+N_i$ and $R$ is a rotation matrix, $T$ is a translation vector, and $N_i$ is instance noise, we want to find $R$ and $T$ that minimize 
\begin{equation}
    \Sigma^2=\sum_{i=1}^N\left\|p_i^{\prime}-\left(R p_i+T\right)\right\|^2.
\end{equation}
Here, we utilize a fast and robust method \cite{arun1987least} to compute the rigid transform matrix, which enables us to map the MR to its paired CT and crop the latter. Due to the potential inaccuracy of the rigid transform, we dilate the cropping range to avoid the risk of over-cropping. After that, the two scans have similar FOVs, so we can use the widely-used DEEDS algorithm~\cite{heinrich2013mrf} to register to the two scans. The obtained deformation field will be applied on the cropping region of the uncropped CT.

\subsection{Cross-modality SAM}
Now, we can train a cross-modality SAM using the registered CT and MR pairs. To accomplish this, we randomly select positive pairs from  the registered regions. Following the original SAM design, for the fine-level embedding learning, we select
$N_{\text{pos}}^f$ positive pairs from the registered regions with overlapping areas and choose $N_{\text{neg}}^f$ points to act as negative samples. Additionally, we choose $N_{\text{fov}}^f$ points from the non-overlapping area of the scan with large FOV. The loss function is defined as follows:
\begin{equation}
\mathcal{L}=-\sum_{i=1}^{N_{\text{pos}}^f} \log \frac{\exp \left(\boldsymbol{x}_i \cdot \boldsymbol{x}_i' / \tau\right)}{\exp \left(\boldsymbol{x}_i \cdot \boldsymbol{x}_i' / \tau\right)+\sum_{j=1}^{N_{\text{neg}}^f} \exp \left(\boldsymbol{x}_i \cdot \boldsymbol{x}_j / \tau\right)+\sum_{j=1}^{N_{\text{fov}}^f} \exp \left(\boldsymbol{x}_i \cdot \boldsymbol{x}_j / \tau\right)},
\end{equation}
where $x_i$ and $x_i'$ are the embeddings of a positive pair, $x_j$ represents the negative sample, and $\tau$ is the temperature parameter. Similarly, for the coarse-level learning, we select $N_{\text{pos}}^c$ positive pairs and $N_{\text{neg}}^c$ negative samples for each positive pair. To fully utilize the data, we also include augmented intra-modality data as input and train using the self-supervised SAM method.

After training the first cross-modality SAM, we can discard the augmented SAM and perform iterative refinement using AdaReg and cross-modality SAM learning. During each iteration, we decrease the margin of the dilation mask, resulting in a closer FOV for the cropped pairs. This leads to a more accurate deformable fine-tuning process. The model will converge after several iterations. We use the converged cross-modality SAM as our final model.

\section{Experiments}
\subsection{Datasets, Metrics and Implementation Details}
We trained and tested our method using two datasets of different body parts: a head-and-neck dataset and an abdomen dataset. The head-and-neck dataset consists of 120 paired T1 MRI and non-contrast CT images that were not registered, with the MRI having a spacing of 0.5$\times$0.5$\times$6 mm and the CT having a spacing of 1$\times$1$\times$3 mm. The T1 MRI images have a limited FOV and mainly capture the region between the nose and the second cervical vertebra, while the CT images include regions from the top of the head to the middle of the lungs. We used 100 cases for training and 20 for testing. The abdomen dataset contains 98 pairs of T2 MRI and non-contrast CT scans, with the CT and MRI having spacings of 0.7$\times$0.7$\times$5 mm and 0.8$\times$0.8$\times$8 mm, respectively. The MRI images have a small FOV around the liver and kidney, while the CT images encompass regions from the chest to the bottom of the pelvis. We used 80 cases for training and 18 cases for testing.

\begin{figure}[tp]
    \centering
    \includegraphics[scale=0.38]{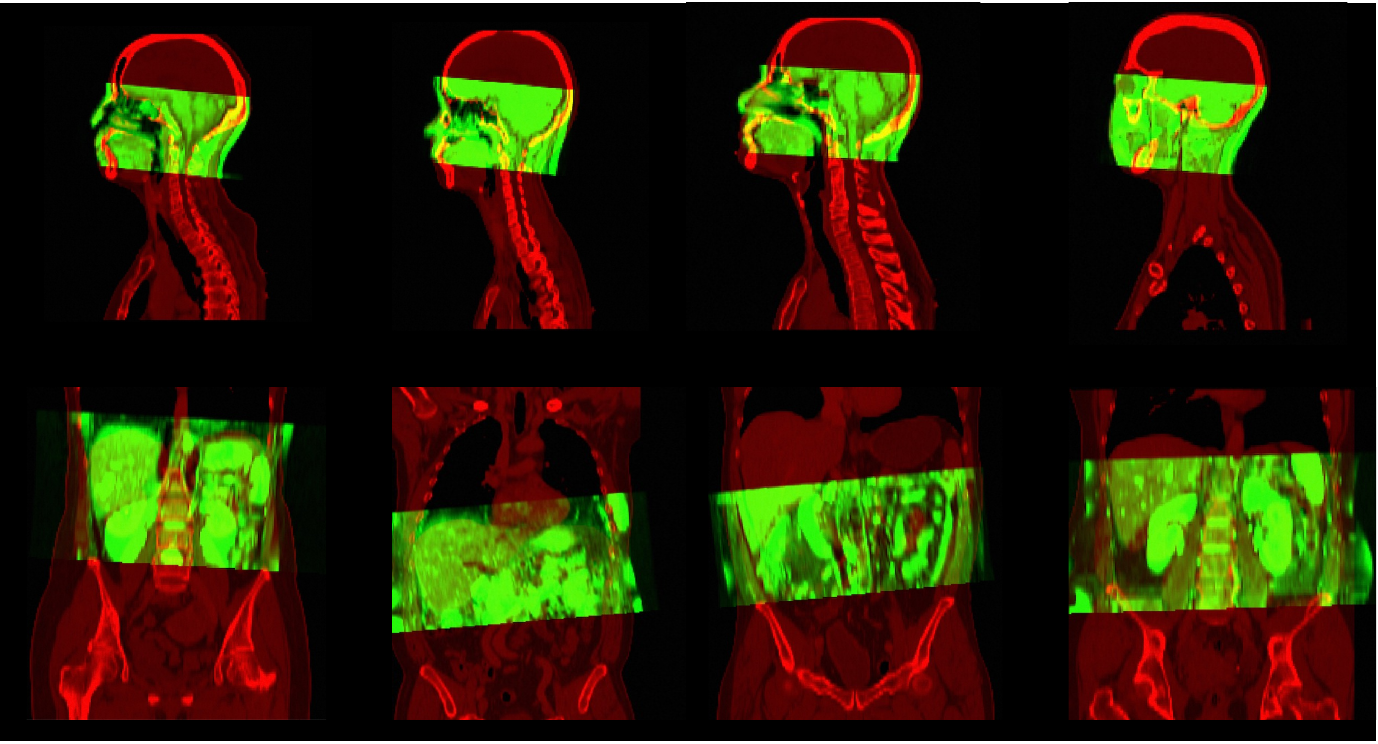}
    \caption{Affine registration results using our method on both head-and-neck and abdomen images. The CT images are represented in red color, while the MRIs are displayed in green.}
    \label{fig:example}
\end{figure}

We assessed the performance of cross-modality affine registration by comparing the mean Euclidean distance (MED) of the same landmarks on the registered pairs. Specifically, we use cross-modality SAM to find correspondences between CT and MR to estimate the affine transform matrix, and then compute the MED of landmarks between the CT and registered MR. We annotated 12 landmarks on both CT and MRI images for the head-and-neck dataset, including the lacrimal gland, the endpoint of the temporomandibular joint, the top and bottom of the C2 spine, the middle point of the jawbone, and the intersection of the lateral pterygoid muscle and upper jawbone. For the abdomen dataset, we annotated 6 landmarks, including the top and bottom points of the liver and spleen, as well as the top points of the kidneys.

We implement our model using PyTorch v1.9. Our model utilized the 3D ResNet-18 backbone and 3D Feature Pyramid Network (FPN) to generate embeddings, following the original SAM method. We optimized the network using Stochastic Gradient Descent (SGD) with a momentum of 0.9 and set the learning rate to 0.02. Prior to training, all scans were resampled to an isotropic resolution of 2mm. During training of the cross-modality SAM, we iterated in each mini-batch with self-supervised learning and registered pair learning. For the self-supervised part, we cropped patches to a size of 96 x 96 x 32 and used a batch size of 4. For registered pairs, we inputted the entire overlapped region along with a randomly selected region in the non-overlapped area. Due to limitations of GPU memory, we only input one pair of data in each mini-batch. For fine-level embedding learning, we selected $N_{\text{pos}}^f=200$ positive pairs and for each positive pair, we randomly selected $N_{\text{neg}}^f=500$ and $N_{\text{fov}}^f=100$ samples from the non-overlapped region.
For coarse-level  we set $N_{\text{pos}}^c=100$ and  $N_{\text{neg}}^c=200$. For the head-and-neck dataset, we set the radius of the dilation structure element to 5 voxels on the first iteration and then reduced it to 2 for the next iteration in the AdaReg module. Only one iteration is used. For the abdomen dataset, we set the radius initially to 15 voxels and reduced it by 5 voxels on each iteration. We trained 3 iterations on the abdomen dataset. Note that although we need several iterations in training, our inference process is one-pass and very fast.

\subsection{Main Results}

Table \ref{tab1} shows the results on the head-and-neck dataset. We compared our method with two widely used conventional methods, DEEDS-affine \cite{heinrich2013mrf} and Elastix \cite{klein2009elastix}, as well as with C2FViT \cite{mok2022affine}, one of the best performance learning-based methods. In the C2FViT method, the local normalized cross-correlation (NCC) loss is used for intra-modality registration, which is not suitable for cross-modality scenarios. Therefore, we replaced it with the local mutual information loss \cite{maes2003medical,chen2022transmorph}. Our method outperformed all other methods by a large margin, as shown in the results.

\begin{table}[t]
\caption{Landmark matching distance (voxels) on the head-and-neck dataset.} \label{tab1}
\resizebox{\textwidth}{!}{
\begin{tabular}{l|c|c|c|c}
\hline
Method  &MED$_{X}$  & MED$_{Y}$ & MED$_{Z}$  &MED  \\
\hline
Initial &1.57$\pm$1.80
&8.66$\pm$4.96
&30.89$\pm$10.38
&32.49$\pm$10.56
\\\hline
DEEDS-Affine \cite{heinrich2013mrf}&12.66$\pm$8.35&13.91$\pm$8.95&17.57$\pm$10.09&28.45$\pm$10.18\\\hline
Simple-elastix \cite{klein2009elastix}&0.95$\pm$0.95&3.79$\pm$4.52&12.20$\pm$19.74&13.40$\pm$19.89\\\hline
C2FViT \cite{mok2022affine}&1.25$\pm$1.03
&2.69$\pm$1.99
&7.92$\pm$4.16
&8.83$\pm$3.99
\\\hline
Ours&\textbf{0.87$\pm$0.83
}&\textbf{1.89$\pm$1.66
}&\textbf{2.29$\pm$1.74
}&\textbf{3.41$\pm$2.11
}\\
\hline
\end{tabular}}
\end{table}
\begin{table}
\caption{Landmark matching distance (voxels) on the abdomen dataset.} \label{tab2}
\resizebox{\textwidth}{!}{
\begin{tabular}{l|c|c|c|c}
\hline
Method  &MED$_{X}$  & MED$_{Y}$ & MED$_{Z}$  &MED  \\
\hline
Initial &11.99$\pm$11.68
&18.14$\pm$16.96
&64.90$\pm$25.62
&72.59$\pm$22.13
\\\hline
DEEDS-Affine \cite{heinrich2013mrf}&\multicolumn{4}{c}{Failed on most cases}\\\hline
Simple-elastix \cite{klein2009elastix}& \multicolumn{4}{c}{Failed on most cases} \\\hline
C2FViT \cite{mok2022affine}&14.59$\pm$11.30
&16.93$\pm$14.75
&22.39$\pm$16.70
&34.37$\pm$17.37
\\\hline
Ours&\textbf{3.63$\pm$2.86
}&\textbf{6.16$\pm$5.69
}&\textbf{8.70$\pm$6.40
}&\textbf{12.86$\pm$6.54
}\\
\hline
\end{tabular}
}
\end{table}

We then tested our method on the more challenging abdomen dataset. Organs such as the stomach and intestine can undergo significant shape changes between paired scans, which can affect the positioning of other organs. Thus, a simple rigid registration may not align the organs perfectly. Moreover, MRI scans may only capture a portion of the organ, adding more difficulty to the task. As shown in Table \ref{tab2}, conventional methods fail to handle this condition and perform poorly in most cases. The C2FViT method reduces the MED from 72.59 to 34.37. Our method achieves even better results by reducing the MED to less than 13 voxels. Our method can provide a good initial alignment to subsequent deformable registration methods in this extremely challenging cross-modality diverse-FOV scenario. We also evaluate the performance of the Nasopharyngeal Carcinoma (NPC) segmentation accuracy using our method as initial affine step, the results are shown in the supplementary. Table \ref{tab3} shows the MED changes at each iteration.
\vspace{-10pt}
\begin{table}
\caption{The MED at each iteration on the abdomen dataset.} \label{tab3}
\resizebox{\textwidth}{!}{
\begin{tabular}{c|c|c|c|c}
\hline
~~~~~Iterations~~~~~  &MED$_{X}$  & MED$_{Y}$ & MED$_{Z}$  &MED  \\
\hline
0 &3.91$\pm$2.98
&7.53$\pm$7.25
&13.46$\pm$9.48
&17.94$\pm$9.05
\\\hline
1 &3.75$\pm$3.35&6.60$\pm$6.07&10.55$\pm$8.49&14.75$\pm$8.44\\\hline
2 &3.63$\pm$2.86&6.16$\pm$5.69&8.70$\pm$6.40&12.86$\pm$6.54\\\hline
\end{tabular}
}
\end{table}
\vspace{-15pt}
\section{Conclusion}
\vspace{-5pt}
We propose Cross-SAM, a framework to learn anatomical embeddings for multi-modality images. Our framework can be used for affine/rigid registration in extremely challenging cross-modality diverse-FOV scenario, providing a good initial alignment to subsequent deformable registration methods.
%
%
%
\bibliographystyle{splncs04}
\bibliography{ijcai22}
\newpage
\section{Appendix}

\vspace{-20pt}
\begin{figure}[h]
    \centering
    \includegraphics[scale=0.32]{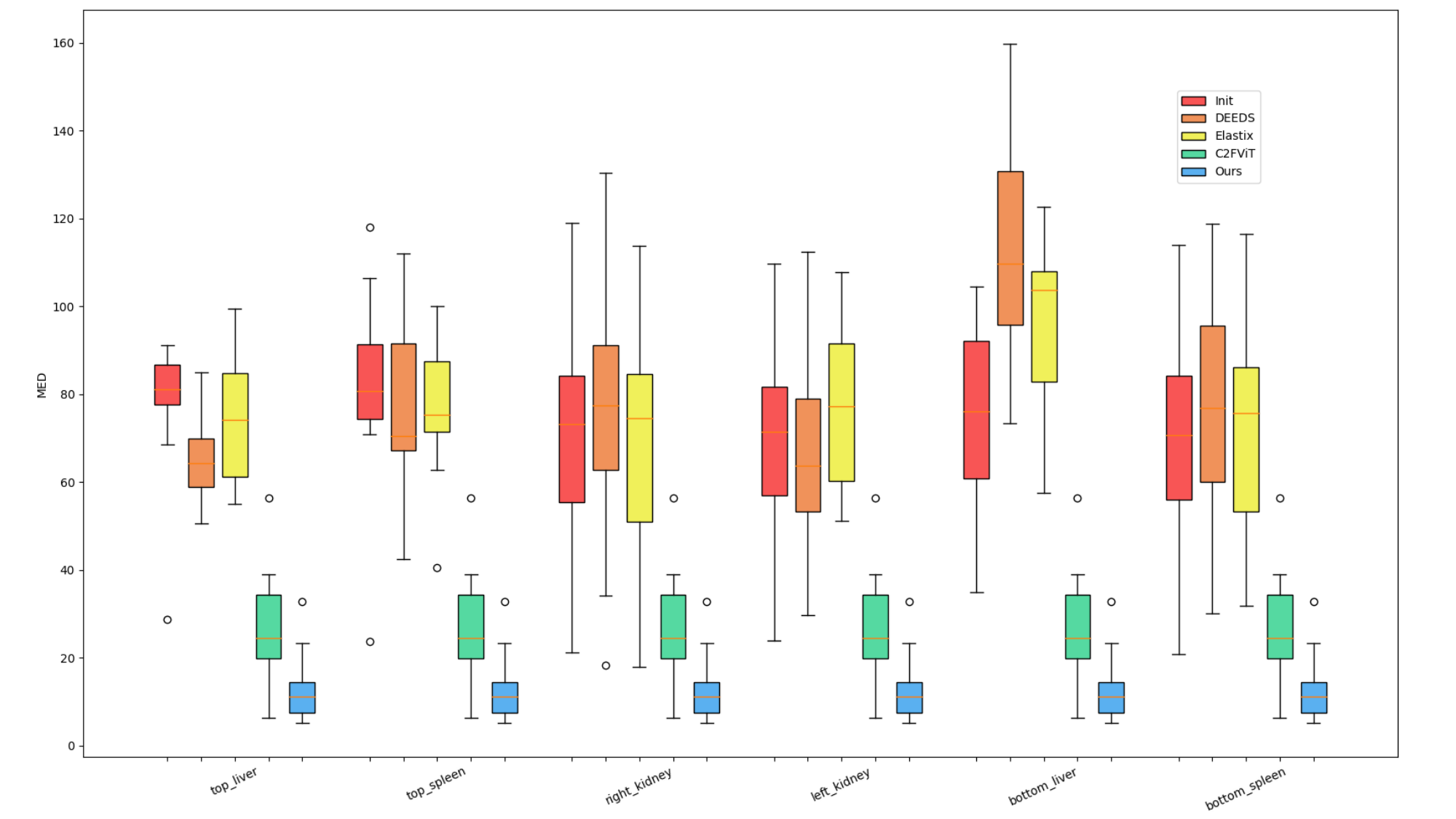}
    \vspace{-10pt}
    \caption{Comparison of the affine registration results on abdomen dataset. Smaller mean Euclidean distance (MED) means better registration accuracy. As we can see DEEDS and Elastix methods failed to reduce the MED. In contrast, our method significantly reduced the MED and outperformed all other methods.}
    \label{fig:abd}
\end{figure}
\vspace{-30pt}
\begin{figure}[h]
    \centering
    \includegraphics[scale=0.38]{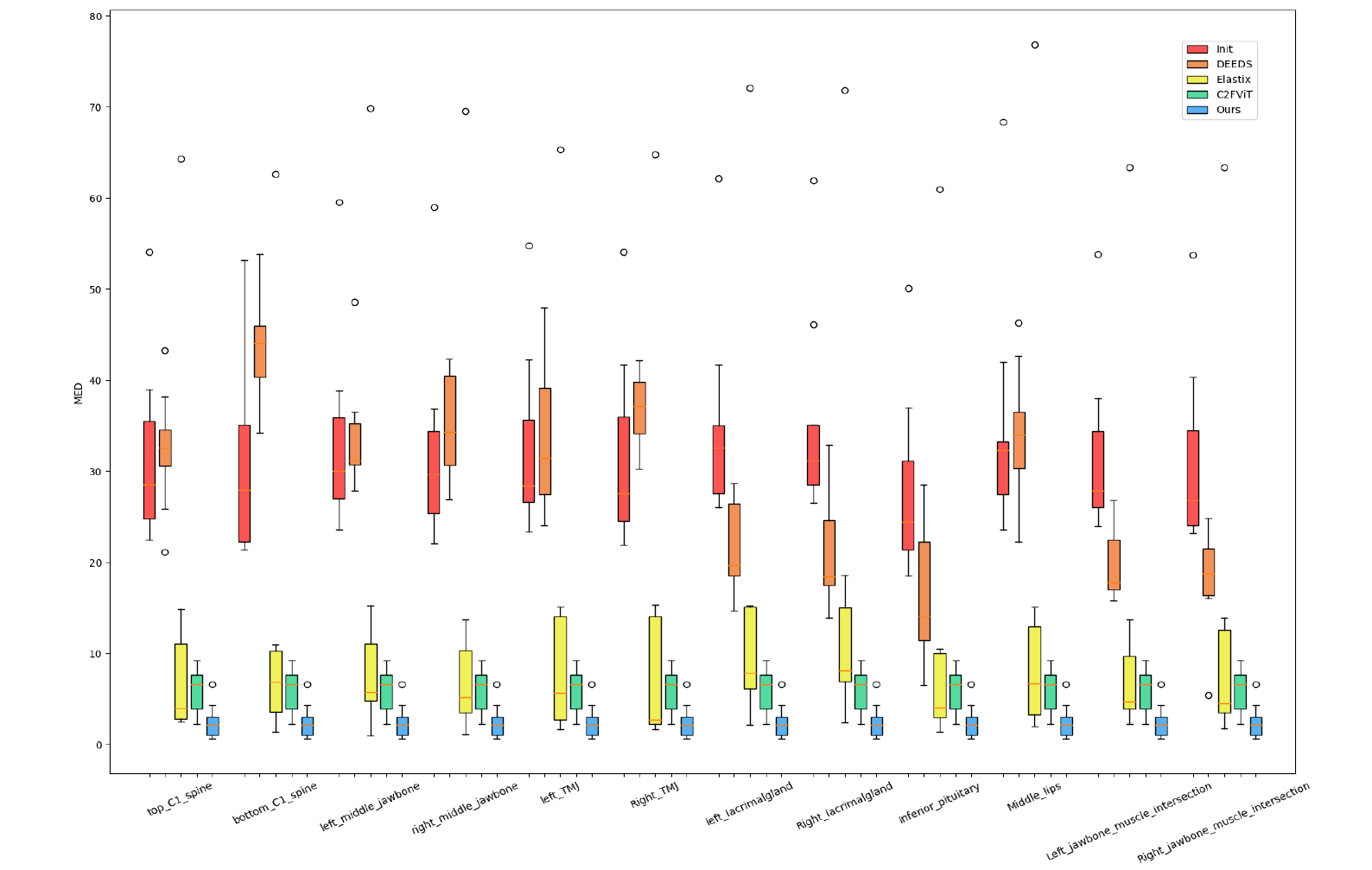}
    \vspace{-10pt}
    \caption{Comparison of the affine registration results on head-and-neck dataset. Our method outperformed all other methods.}
    \label{fig:npc}
\end{figure}
\begin{figure}[h]
    \centering
    \includegraphics[scale=0.30]{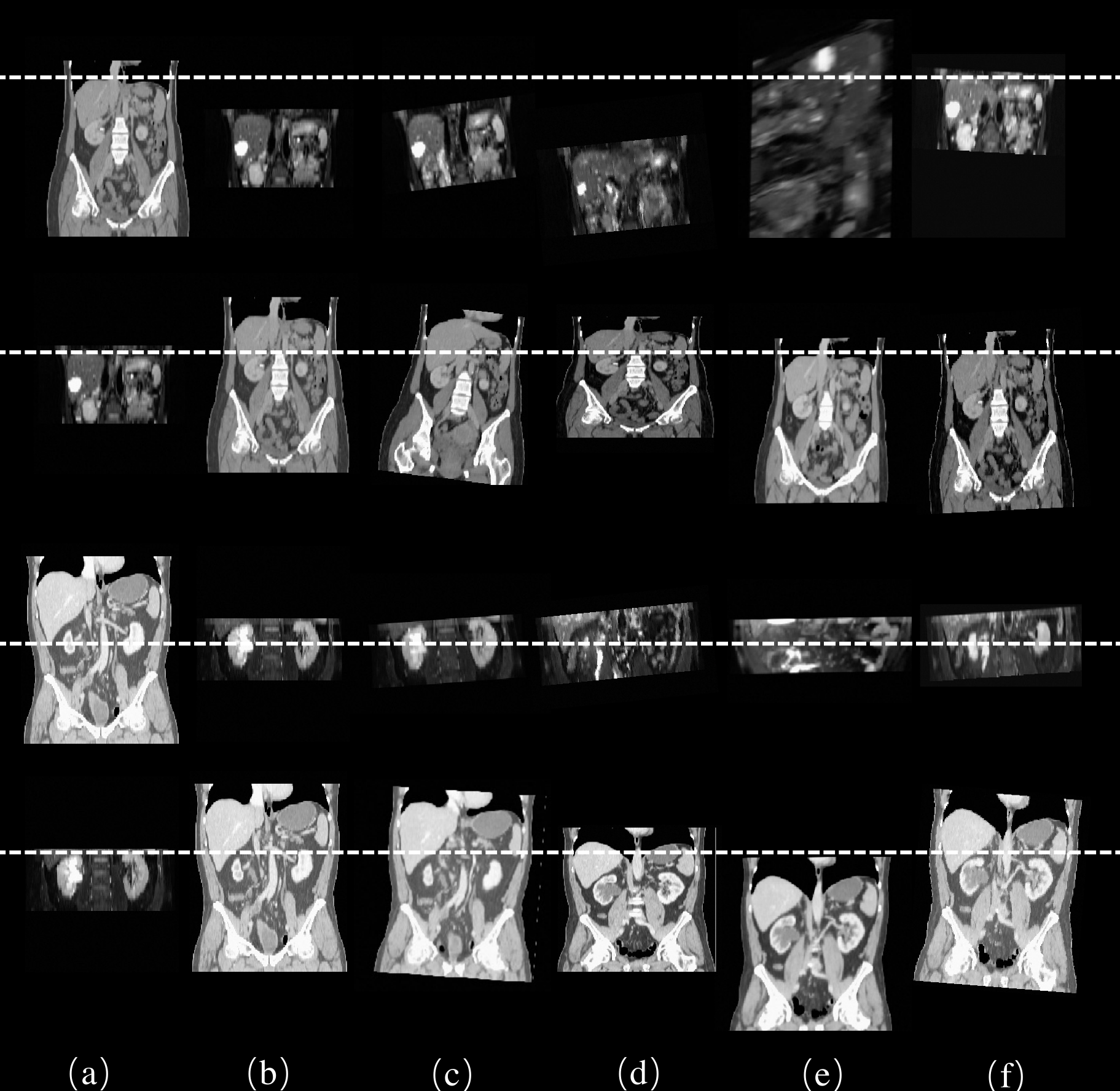}
    \caption{Visualization of using different methods for affine registration on the abdomen dataset. (a) Fixed images, (b) moving images, (c) DEEDS-affine results, (d) Elastix-affine results, (e) C2FViT results, and (f) our own results. We show both CT to MRI (large FOV to small FOV) and MRI to CT (small FOV to large FOV) registration.}
    \label{fig:example}
\end{figure}

\begin{table}[h]
\caption{Comparison results on nasopharyngeal carcinoma gross tumor volume (GTV) segmentation.} \label{tab1}
\resizebox{\textwidth}{!}{
\begin{tabular}{c|c|c|c|c}
\hline
Method  &HD95  & Precision & Recall  &Dice  \\
\hline
CT-only &8.831$\pm$5.717
&0.764$\pm$0.163
&\textbf{0.776$\pm$0.156}
&0.741$\pm$0.085
\\\hline
CT-MRI manual-crop + DEEDS &8.558$\pm$4.893&0.792$\pm$0.154&0.759$\pm$0.146&0.750$\pm$0.082
\\\hline
CT-MRI CrossSAM + DEEDS&\textbf{7.757$\pm$3.691
}&\textbf{0.830$\pm$0.138
}&0.747$\pm$0.140
&\textbf{ 0.764$\pm$0.081
}\\
\hline
\end{tabular}}
\vspace{-10pt}
\caption*{\small *CT-only means using only manually cropped CT to do segmentation. CT-MRI manual-crop means manually cropping the CT volume to the rough region between the nose and the second cervical vertebra, followed by registering MRI to the CT volume. Our CT-MRI CrossSAM method involves adaptive cropping of the region based on the field-of-view (FOV) of the MRI volume. For all three methods, we used DEEDS for deformable registration and nnUNet model for segmentation.}
\end{table}

%




\end{document}